\pdfoutput=1

\documentclass[11pt]{article}

\usepackage{EMNLP2023}

\usepackage{times}
\usepackage{latexsym}
\usepackage{smartdiagram}
\usepackage{tikz}
\usepackage{tcolorbox}
\usetikzlibrary{arrows.meta,
                positioning
                }
\usetikzlibrary{positioning, shapes}
\usepackage{threeparttable}
\usepackage{graphicx}
\usepackage[T1]{fontenc}

\usepackage[utf8]{inputenc}
\usepackage{enumitem}
\usepackage{microtype}
\usepackage{amsmath}
\DeclareMathOperator{\abs}{\mathsf{abs}}
\renewcommand{\max}{\mathsf{max}}
\newcommand{\mfun}[1]{$\mathsf{#1}$}
\newcommand{\qaid}[1]{
\centering{
\begin{tcolorbox}[colback=orange!20, colframe=white, left=0pt, right=0pt, top=0pt, bottom=0pt, width=0.3\textwidth]
\centering{\tn{#1}}
\end{tcolorbox}}
}
\newcommand{\qabox}[2]{
\centering{
\begin{tcolorbox}[colback=#1!10, colframe=white, left=0pt, right=0pt, top=0pt, bottom=0pt]
#2
\end{tcolorbox}}
}
\newcommand{\gptbox}[2]{
\centering{
\begin{tcolorbox}[colback=#1!7, colframe=white, left=0pt, right=0pt, top=0pt, bottom=0pt]
{\color{#1} #2}
\end{tcolorbox}}
}
\usepackage{inconsolata}
\usepackage{xcolor}
\usepackage{amsthm, amsmath, amsfonts}
\usepackage{bm}
\definecolor{deepblue}{HTML}{1f77b4}
\definecolor{vibrantorange}{HTML}{ff7f0e}
\definecolor{lushgreen}{HTML}{2ca02c}
\definecolor{elegantpurple}{HTML}{9467bd}
\definecolor{olivegreen}{RGB}{128, 128, 0}
\definecolor{darkgreen}{RGB}{0, 100, 0}

\newcommand{\clb}[1]{{\color{blue} #1}}

\newcommand{\cldg}[1]{{\color{darkgreen} #1}}
\newcommand{\clm}[1]{{\color{magenta} #1}}
\newcommand{\clo}[1]{{\color{orange} #1}}

\tikzset{
  heading/.style={
    rectangle,
    rounded corners,
    draw=gray,
    align=center,
    minimum width=8em,
    text width=8em,
    font=\scriptsize,
    fill=orange!70!yellow!30,
    top color=orange!70!yellow!50,
    bottom color=orange!70!yellow!10,
    drop shadow={shadow xshift=2pt, shadow yshift=-2pt, fill=gray!70, opacity=0.6}
  },
  example_heading/.style={
    rectangle,
    rounded corners,
    draw=gray,
    align=center,
    minimum width=8em,
    text width=10em,
    font=\scriptsize,
  },
  example_prompt/.style={
    rectangle,
    rounded corners,
    draw=gray,
    align=center,
    minimum width=8em,
    text width=40em,
    font=\scriptsize,
  },
  inline_prompt/.style={
    rectangle,
    rounded corners,
    draw=gray,
    align=left,
    minimum width=18em,
    text width=18em,
    font=\scriptsize,
  },
  prompt/.style={
    rectangle,
    rounded corners,
    draw=gray,
    align=left,
    minimum width=18em,
    text width=18em,
    font=\scriptsize,
    fill=cyan!70!blue!20,
    top color=cyan!70!blue!30,
    bottom color=cyan!70!blue!10,
    drop shadow={shadow xshift=2pt, shadow yshift=-2pt, fill=gray!70, opacity=0.6}
  },
  response/.style={
    rectangle,
    rounded corners,
    draw=gray,
    align=center,
    minimum width=18em,
    text width=18em,
    font=\scriptsize,
    fill=magenta!50!red!20,
    top color=magenta!50!red!30,
    bottom color=magenta!50!red!10,
    drop shadow={shadow xshift=2pt, shadow yshift=-2pt, fill=gray!70, opacity=0.6}
  },
  arrow/.style={
    ->,
    draw=gray,
    line width=2mm,
    shorten >=1pt,
    shorten <=1pt,
    -{Triangle[scale=0.6]}
  }
}

\newcommand\tn[1]{\textbf{\small #1}}
\title{Zero-Shot Question Answering over Financial Documents using Large Language Models}

\author{
\bf{Karmvir Singh Phogat,} 
\bf{Chetan Harsha,}  
\bf{Sridhar Dasaratha,}\\ 
\bf{Shashishekar Ramakrishna,}
\bf{Sai Akhil Puranam}
\\
EY Global Delivery Services India LLP
\\
\texttt{\{Karmvir.Phogat, Chetan.Harsha, Sridhar.Dasaratha\}@gds.ey.com},\\
\texttt{\{Shashishekar.R, Sai.Puranam\}@gds.ey.com}
}

\begin{document}
\maketitle
\begin{abstract}
We introduce a large language model (LLM) based approach to answer complex questions requiring multi-hop numerical reasoning over financial reports. While LLMs have exhibited remarkable performance on various natural language and reasoning tasks, complex reasoning problems often rely on few-shot prompts that require carefully crafted examples. In contrast, our approach uses novel zero-shot prompts that guide the LLM to encode the required reasoning into a Python program or a domain specific language. The generated program is then executed by a program interpreter, thus mitigating the limitations of LLM in performing accurate arithmetic calculations.

We evaluate the proposed approach on three financial datasets using some of the recently developed generative pretrained transformer (GPT) models and perform comparisons with various zero-shot baselines. The experimental results demonstrate that our approach significantly improves the accuracy for all the LLMs over their respective baselines. We provide a detailed analysis of the results, generating insights to support our findings. The success of our approach demonstrates the enormous potential to extract complex domain specific numerical reasoning by designing zero-shot prompts to effectively exploit the knowledge embedded in LLMs.
\end{abstract}

\section{Introduction}
In recent years, the development of large language models (LLMs) has achieved significant advances in natural language processing (NLP). Typically, LLMs are pretrained on large corpora of text from the internet which has given rise to the capability of adapting to a wide variety of new tasks from different domains without the need for huge amount of task specific data. Scaling up the size of these models has not only improved sampling efficiency and performance, \cite{scaling_laws} but also introduced reasoning capabilities \cite{emergent_abilities,cot,zs_cot}. 

LLMs have been shown to perform well on tasks requiring reasoning capabilities in various domains, including code writing \cite{evaluating_llm_code}, math problem solving \cite{solving_qr,formal_mathematics}, dialogue \cite{improving_alignment,lamda}, common sense reasoning \cite{unsupervised_commonsense,palm} and symbolic reasoning \cite{cot,sc_cot}. The design of the prompt, known as prompt engineering, plays a significant role in adapting the pretrained LLMs to new tasks with little or no task specific training data. Recently, there has been extensive work \cite{pre_train} which demonstrates the importance of prompt design in usage of the LLMs and unlocking their reasoning capabilities. However, \cite{dissociating_language} argue that LLMs cannot combine elementary knowledge with common sense reasoning. \cite{llm_plan} claim that benchmarks on which LLMs show reasoning capabilities are simplistic and cannot be used as evidence. \cite{sparks, multi_task} show that LLMs face challenges in numerical reasoning. Hence, adapting LLMs to new domains requires prompt engineering and a system design that can overcome the limitations of LLMs. 

Question answering in the financial domain is an active area of research which could potentially benefit from the use of LLMs with appropriate system design. Financial question answering involves numerous steps and complex numerical reasoning with precise arithmetic calculations, making it more challenging than classical question answering problems \cite{hotpotqa,know_what}. Typically for complex problems, few-shot prompt based approaches have been used \cite{cot,pot}. However it has been shown that the output of the LLMs is sensitive to the few-shot samples used as well as to the ordering of those samples \cite{fantastically}. Further, the samples can contain large number of tokens and providing multiple samples for few-shot prompts would increase the number of input tokens, sometimes even crossing the limit of LLMs. Hence,  designing and using few-shot prompts for financial question answering can become quite challenging. 

We propose a new approach using zero-shot prompts for financial question answering with LLMs, thus eliminating the requirement to create hand crafted examples. These prompts contain high-level instructions to guide the encoding of financial reasoning process into a Python program (ZS-FinPYT) or a domain specific language (ZS-FinDSL). For ZS-FinPYT, we achieve the zero-shot system by instructions that layout the high-level approach to generate a valid Python program, while for ZS-FinDSL we enable the same by identifying a program structure for robust domain-specific languages (DSL) program extraction. In both cases, the generated program is executed externally by a program executor to provide the final answer. 
We evaluate the use of the latest GPT-x models on their ability to perform financial reasoning as they have shown state-of-the-art performance on various tasks involving question answering and reasoning \cite{gpt4_report,mathematical_chatgpt,chatgpt_ai}. Specifically, we explore the use of the GPT models text-davinci-003, gpt-3.5-turbo and gpt-4 in answering financial questions.

We evaluate the proposed approach on three financial question answering datasets, with three different GPT models and compare with various baselines. The experimental results demonstrate that our approach significantly improves the accuracy for all models. The success of our approach demonstrates the enormous potential to extract complex domain specific numerical reasoning by carefully designing LLM based systems for specific applications and crafting prompts to effectively exploit the knowledge embedded in the LLMs.

\section{Background}
NLP techniques have proven useful to solve various problems in the financial domain such as sentiment analysis to assist market prediction \cite{dl_financial,multilayer} and fraud detection for risk management \cite{next_gen_aml,detecting_fraud}. Financial domain specific language models have been trained on large scale financial data and fine tuned for specific problems \cite{finbert}. \cite{finqa} introduce a large-scale question answering dataset, FinQA and propose FinQANet with a retriever-generator architecture based on pretrained BERT like models. 

With the introduction of LLMs, it has become feasible to directly use these language models without domain specific pretraining. \cite{convfinqa} propose a large-scale financial dataset, ConvFinQA for conversational question answering. They propose a few-shot prompt (with 16 exemplars) based approach using GPT-3 text-davinci-002 model to generate a DSL program.

One of the key techniques which significantly improves reasoning abilities of LLMs is chain-of-thought prompting introduced by \cite{cot}. They propose a few-shot prompt that consists of triples: <input, chain-of-thought, output>, where the chain-of-thought (CoT) is a series of intermediate natural language reasoning steps that leads to the final output. \cite{zs_cot} demonstrate that reasonable zero-shot learning is achieved by simply adding “Let’s think step by step” to the prompt and using a two-prompt approach: the first prompt to extract the reasoning path and the second to extract the final answer. Unlike our approach, which avoids performing calculations using the LLM, both of these approaches utilize the LLM for generating mathematical expressions that encode the reasoning and perform arithmetic at each step.

Program of thoughts (PoT) prompting \cite{pot} and Program-aided Language Models (PAL) \cite{pal} are approaches that are conceptually similar to our proposed technique. However, \cite{pot} show only limited zero-shot prompting experiments for financial data sets.  Their results indicate that few-shot prompting significantly outperforms the zero-shot prompts. \cite{pal} discuss only few-shot prompting and do not show any results on financial data sets. In contrast, our work focuses entirely on optimizing zero-shot prompts that generate Python program or domain specific language for financial question answering. We further demonstrate that carefully designed zero-shot prompts for financial question answering can achieve comparable results with few-shot methods. 

\section{Zero-shot Prompting for Financial Domains}
We introduce a novel zero-shot template-based prompting for financial question answering. These prompts are designed to generate executable programs for answering questions. The executable program generation and their execution enables accurate mathematical calculations which eliminates arithmetic errors. We follow the prompt guidelines described in \cite{reynolds} and employ the following principles for designing zero-shot prompting for question answering: 

\textbf{Signifier:} A signifier is a pattern which keys the intended behavior. A task specific signifier directly elucidates the task at hand. The sentence -- ``Read the following passage and then answer the question'', specifically describes the question answering task that is to be performed. 

\textbf{Memetic proxy:} A memetic proxy is a concept in which a character or characteristic situation is used as a proxy for an intention. ``\#Python'' can be a memetic proxy for the LLM to clarify the intention that the response should have a Python program. 

\textbf{Constraining behavior:} In addition to directing the LLM on the desirable response, it is important for the prompt to inform the LLM of undesirable responses. Instructions restricting undesirable LLM responses fall into the constraining behavior category.    

\textbf{Meta prompting:} A meta prompt is a short phrase or a fill-in-the-blank template encapsulating a more general intention that will unfold into a more specific prompt when combined with additional information such as the task at hand. In the question answering task, the sentence --``Let us think step by step.'', elicits step-by-step reasoning in LLMs for answering questions.     

Inspired by these prompt design principles, we present two zero-shot prompting techniques: ZS-FinPYT prompt that enables LLMs to generate Python executable programs and ZS-FinDSL prompt that enables LLMs to generate executable domain specific language programs. We also discuss two baseline zero-shot prompting techniques, one using a simple dual prompt and another using zero-shot chain-of-thought prompting (ZS-CoT) motivated by \cite{zs_cot}. For reproducibility purposes, we provide exact prompts for all techniques and datasets. 

\subsection{Zero-shot FinPYT}
The ZS-FinPYT prompt is a collection of instructions that directs the LLM to generate a valid Python program that can be executed by the \mfun{exec} function. Based on preliminary experiments, we identified the following requirements for the proposed prompt: 
\begin{enumerate}[label=(\alph*), nosep, align=left, leftmargin=*]
\item The prompt should describe the task such that it enables the LLM to generate consistent programs for answering the questions.           
\item The final answer to a question must be stored in a specified Python variable for enabling consistent  extraction of executed answer.  
\item The LLM generated program should not include Python non-executable statements for seamless execution of Python programs. 
\end{enumerate}

The ZS-FinPYT prompt is designed to accommodate the above requirements in the following manner:

\textbf{Direct task specification using the signifier:} 
We use the following signifier for explicitly specifying the question answering task: 
\begin{center}
\begin{tikzpicture}[node distance=12mm]
	\node (n1) [inline_prompt] {Read the following passage and then write Python code to answer the question:\\ 
	\textbf{Passage:} \clb{\texttt{text} + \texttt{table}} \\
	\textbf{Question:} \clb{\texttt{ask question?}} \\
	Answer this question by following the below instructions.
}; 
\end{tikzpicture}
\end{center}
The signifier explicitly calls out the task of writing a Python program to answer the question after reading the passage where the passage and the questions are identified with the identifiers ``\textbf{Passage:}'' and ``\textbf{Question:}'' respectively. Furthermore, the prompt directs the LLM to follow certain instructions while answering the question.  

\textbf{Direct sub-task specification using the signifier:}
The sub-task of storing the final answer to a specific Python variable is described as a part of instructions to the LLM:  
\begin{center}
\begin{tikzpicture}[node distance=12mm]
	\node (n1) [inline_prompt] {Define the Python variable which must begin with a character. \\Assign values to variables required for the calculation. \\Create Python variable "ans" and assign the final answer (bool/float) to the variable "ans".
}; 
\end{tikzpicture}
\end{center}

\textbf{Constraining LLM behavior:} 
To ensure naming conventions are followed and prevent the generation of non-executable statements, we include the following instructions in the prompt:
\begin{center}
\begin{tikzpicture}[node distance=12mm]
	\node (n1) [inline_prompt] {Define the Python variable which must begin with a character. Don't include non-executable statements and include them as part of comments.
}; 
\end{tikzpicture}
\end{center}

\textbf{Memetic proxy phrases:} 
Certain memetic proxy phrases are employed to implicitly convey intentions. For instance, the memetic phrase ``\#Comment: $\ldots$'' guides the LLM to understand that comments are always preceded by the ``\#'' character. Similarly, the memetic phrase ``\#Python'' instructs the LLM to generate a Python program.

\begin{figure}[htb]
	\begin{center}
\begin{tikzpicture}[node distance=12mm]
	\node (n1) [prompt] {Read the following passage and then write Python code to answer the question:\\ 
	\textbf{Passage:} \clb{\texttt{text} + \texttt{table}} \\
	\textbf{Question:} \clb{\texttt{ask question?}} \\
	Answer this question by following the below instructions. \\
	\textbf{Instructions:} \\  
		\phantom{new} \clm{Define the Python variable which must begin with a character. \\}
		\phantom{new} Assign values to variables required for the calculation. \\
		\phantom{new} Create Python variable "ans" and assign the final answer \\
		\phantom{new} (bool/float) to the variable "ans".\\
		\phantom{new} \clm{Don't include non-executable statements and include them as \\
		\phantom{new} part of comments.} \clo{\#Comment: $\ldots$}\\ 
	Python executable code is: \\
	\clo{\textbf{\#Python}}
}; 

\node (n3) [response, below=of n1]
	      {\texttt{Python code from the LLM.}};
\draw [arrow] (n1) --  (n3);
\node[above left=0.25cm and -3.25cm of n3] (image) {\includegraphics[height=0.7cm]{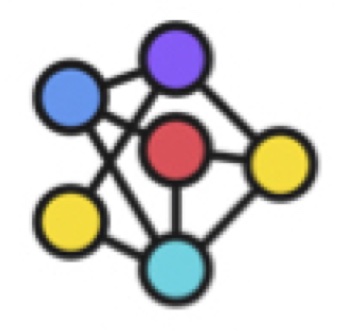}};
\node[below= -0.25cm of image] {\scriptsize LLM};
	\node [fill=black, minimum width=0.05cm, minimum height=0.05cm, node distance=3mm, below left=0.25cm and -0.4cm of n3](c1) {};
	\node [right= -0.05cm of c1](cn1) {\scriptsize Signifier};
	\node [fill=cyan!05!orange, minimum width=0.05cm, minimum height=0.05cm, right= -0.01cm of cn1](c2) {};
	\node [right= -0.05cm of c2](cn2) {\scriptsize \clo{Memetic proxy}};
	\node [fill=cyan!05!magenta, minimum width=0.05cm, minimum height=0.05cm, right= -0.01cm of cn2](c3) {};
	\node [right= -0.05cm of c3](cn3) {\scriptsize \clm{Constraining behavior}};
	\node [fill=blue, minimum width=0.05cm, minimum height=0.05cm, right= -0.01cm of cn3](c4) {};
	\node [right= -0.05cm of c4](cn4) {\scriptsize \clb{Input}};
\end{tikzpicture}
	\end{center}
\caption{ZS-FinPYT prompt for FinQA}
\label{fig:zs_fin_pyt}
\end{figure}
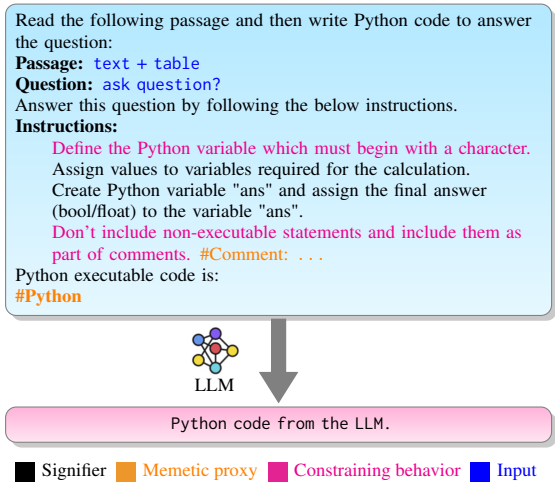

The ZS-FinPYT prompt for the FinQA dataset is depicted in Figure \ref{fig:zs_fin_pyt}. 

\subsection{Zero-shot FinDSL}
The zero-shot FinDSL (ZS-FinDSL) is a zero-shot prompting technique for program generation in a domain specific language (DSL). We use a DSL similar to \cite{finqa} with two differences: we don't have table operators and instead we have a max and min operator. The output of the system is a DSL program that is extracted using a Python script and executed using a language interpreter. In the ZS-FinDSL technique, we adopt a dual prompt approach to extract reasoning for answering questions and generating the corresponding DSL program. 

\subsubsection{Reasoning Extraction Prompt}
The reasoning extraction prompt of ZS-FinDSL consists of two parts: 

\textbf{Direct task specification using the signifier:} 
The question answering task is specified explicitly using the following signifier: 
\begin{center}
\begin{tikzpicture}[node distance=12mm]
	\node (n1) [inline_prompt] {Read the following passage and then answer the question:\\ 
	\textbf{Passage:} \clb{\texttt{text} + \texttt{table}} \\
	\textbf{Question:} \clb{\texttt{ask question?}} \\
}; 
\end{tikzpicture}
\end{center}

\textbf{Meta prompting for reasoning:} 
For generating step by step reasoning for answering the question, the following meta prompt is used: 
\begin{center}
\begin{tikzpicture}[node distance=12mm]
	\node (n1) [inline_prompt] {Answer this question by finding the relevant values and performing step by step calculations. 
}; 
\end{tikzpicture}
\end{center}

\subsubsection{Program Extraction Prompt}
The primary goal of the program extraction prompt is to extract DSL programs from the LLM's response obtained through the reasoning extraction prompt. To achieve this, the program extraction prompt involves specifying the task of program extraction and constraining the LLM's behavior by incorporating domain-specific knowledge.

\textbf{Direct task specification using the signifier:} 
The program extraction task is specified using the following signifier: 
\begin{center}
\begin{tikzpicture}[node distance=12mm]
	\node (n1) [inline_prompt] { 
	\textbf{Question:} \clb{\texttt{ask question?}} \\
	\textbf{Answer:} \clb{\texttt{Answer with reasoning from LLM.}} \\
	\textbf{Task:} From the above question-answer, extract the calculations that were performed to arrive at the answer. The calculations should be provided in the following format:\\
	      \{"\textsc{Program}":\{"\#0":\{\textsc{operation}:"[\texttt{arithmetic/logic}]",\\
	      \phantom{\{"\textsc{Program}":\{"\#0":} \textsc{arg1}:"[\texttt{float/int}]", \textsc{arg2}:"[\texttt{float/int}]"\}, \\
	      \phantom{\{"\textsc{Program}":} "\#1":\{\textsc{operation}:[\texttt{arithmetic/logic}], \\
	      \phantom{\{"\textsc{Program}":\{"\#0":} \textsc{arg1}:"\#0", \textsc{arg2}:"[\texttt{float/int/\#int}]"\}, ...\}, \\
	      \textsc{ "Answer"}: "[\texttt{numerical/boolean}]"\} 
}; 
\end{tikzpicture}
\end{center}

\textbf{Constraining LLM behavior:} 
To ensure consistent program extraction, we limit the mathematical operations to the set specified by the DSL. These operations are commonly used for financial question answering. Moreover, we constrain the program's output to numerical or boolean values to make it executable. The following instructions are passed to the LLM to ensure consistent program extraction:
 
\begin{center}
\begin{tikzpicture}[node distance=12mm]
	\node (n1) [inline_prompt] { 
	      Operation should strictly be restricted to \{add, subtract, multiply, divide, exponent, greater-than, max, min\} only. \\
	      When evaluated the program should only generate numerical or boolean values.
}; 
\end{tikzpicture}
\end{center}
 
\begin{figure*}[htb]
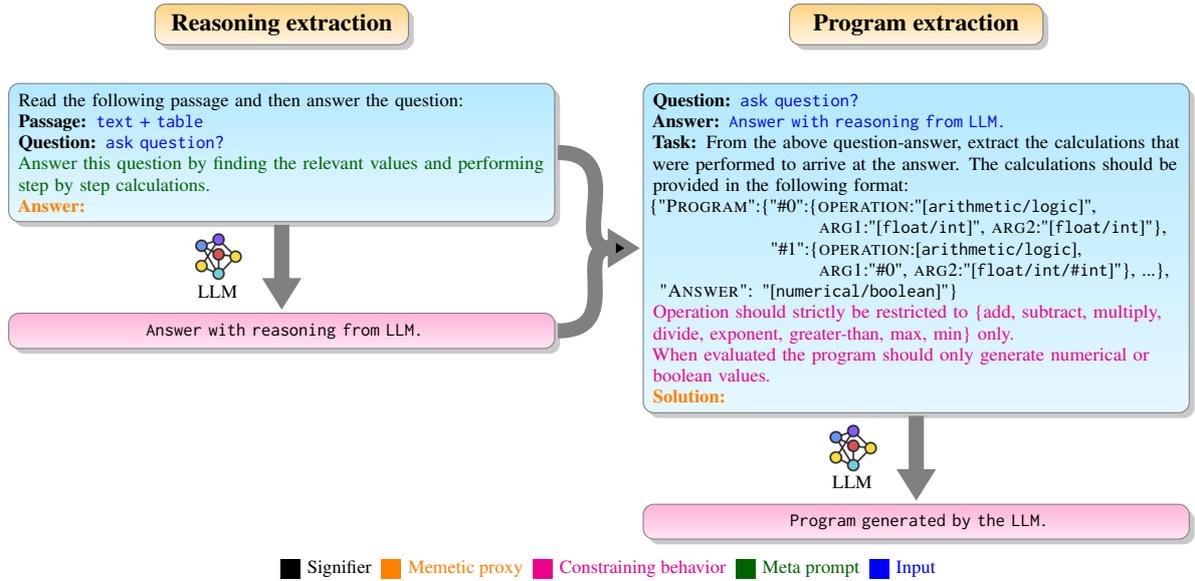

	\begin{center}
\begin{tikzpicture}[node distance=12mm]
	\node (a1) [heading] {\tn{Reasoning extraction}};
	\node (a2) [heading, right= 5cm of a1] {\tn{Program extraction}};
	\node (n1) [prompt, node distance=5mm, below=of a1] {Read the following passage and then answer the question:\\ 
	\textbf{Passage:} \clb{\texttt{text} + \texttt{table}} \\
	\textbf{Question:} \clb{\texttt{ask question?}} \\
	\cldg{Answer this question by finding the relevant values and performing step by step calculations.\\}
	\clo{\textbf{Answer:} }}; 
\node (n2) [prompt, node distance=5mm, below=of a2] 
	      {\textbf{Question:} \clb{\texttt{ask question?}} \\
	      \textbf{Answer:} \clb{\texttt{Answer with reasoning from LLM.}}\\
	      \textbf{Task:} From the above question-answer, extract the calculations that were performed to arrive at the answer. The calculations should be provided in the following format:\\
	      \{"\textsc{Program}":\{"\#0":\{\textsc{operation}:"[\texttt{arithmetic/logic}]",\\
	      \phantom{\{"\textsc{Program}":\{"\#0":} \textsc{arg1}:"[\texttt{float/int}]", \textsc{arg2}:"[\texttt{float/int}]"\}, \\
	      \phantom{\{"\textsc{Program}":} "\#1":\{\textsc{operation}:[\texttt{arithmetic/logic}], \\
	      \phantom{\{"\textsc{Program}":\{"\#0":} \textsc{arg1}:"\#0", \textsc{arg2}:"[\texttt{float/int/\#int}]"\}, ...\}, \\
	      \textsc{ "Answer"}: "[\texttt{numerical/boolean}]"\} \\
	      \clm{Operation should strictly be restricted to \{add, subtract, multiply, divide, exponent, greater-than, max, min\} only. \\
	      When evaluated the program should only generate numerical or boolean values.\\}
	\clo{\textbf{Solution:}} };
	
\node (n4) [response, below=of n2]
              {\texttt{Program generated by the LLM.}};
\node (n3) [response, below=of n1]
	      {\texttt{Answer with reasoning from LLM.}};
\node[above left=0.25cm and -3.25cm of n3] (image1) {\includegraphics[height=0.7cm]{llm.jpg}};
\node[below= -0.25cm of image1] {\scriptsize LLM};
\node[above left=0.25cm and -3.25cm of n4] (image2) {\includegraphics[height=0.7cm]{llm.jpg}};
\node[below= -0.25cm of image2] {\scriptsize LLM};
\draw [arrow] (n1) edge[out=0, in=180, looseness=1.5] (n2);
\draw [arrow] (n3) edge[out=0, in=180, looseness=1.5] (n2);
\draw [arrow] (n1) -- (n3);
\draw [arrow] (n2) -- (n4);
	\node [fill=black, minimum width=0.05cm, minimum height=0.05cm, node distance=3mm, below left=0.25cm and 4.5cm of n4](c1) {};
	\node [right= -0.05cm of c1](cn1) {\scriptsize Signifier};
	\node [fill=orange, minimum width=0.05cm, minimum height=0.05cm, right= -0.01cm of cn1](c2) {};
	\node [right= -0.05cm of c2](cn2) {\scriptsize \clo{Memetic proxy}};
	\node [fill=magenta, minimum width=0.05cm, minimum height=0.05cm, right= -0.01cm of cn2](c3) {};
	\node [right= -0.05cm of c3](cn3) {\scriptsize \clm{Constraining behavior}};
	\node [fill=darkgreen, minimum width=0.05cm, minimum height=0.05cm, right= -0.01cm of cn3](c4) {};
	\node [right= -0.05cm of c4](cn4) {\scriptsize \cldg{Meta prompt}};
	\node [fill=blue, minimum width=0.05cm, minimum height=0.05cm, right= -0.01cm of cn4](c5) {};
	\node [right= -0.05cm of c5](cn5) {\scriptsize \clb{Input}};
\end{tikzpicture}
	\end{center}
\caption{ZS-FinDSL prompt for FinQA}
\label{fig:zs_fin_dsl}
\end{figure*}

The ZS-FinDSL prompt for the FinQA dataset is shown in Figure \ref{fig:zs_fin_dsl}. 

\subsection{Zero-shot Standard Dual}
\begin{figure*}[htb]
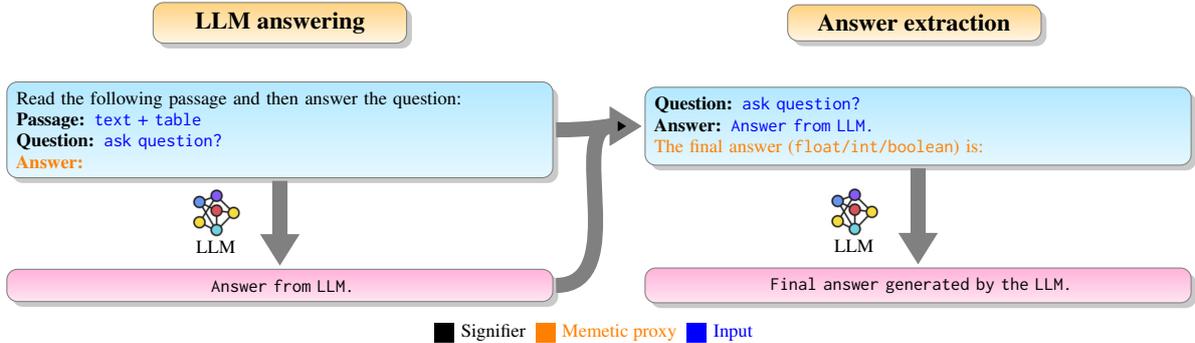

	\begin{center}
\begin{tikzpicture}[node distance=12mm]
	\node (a1) [heading] {\tn{LLM answering}};
	\node (a2) [heading, right= 5cm of a1] {\tn{Answer extraction}};
	\node (n1) [prompt, node distance=5mm, below=of a1] {Read the following passage and then answer the question:\\ 
	\textbf{Passage:} \clb{\texttt{text} + \texttt{table}} \\
	\textbf{Question:} \clb{\texttt{ask question?}} \\
	\clo{\textbf{Answer:}} }; 
\node (n3) [response, below=of n1]
	      {\texttt{Answer from LLM.}};
\node (n4) [response, right=of n3]
              {\texttt{Final answer generated by the LLM.}};
\node (n2) [prompt, node distance=13.5mm, above=of n4] 
	      {\textbf{Question:} \clb{\texttt{ask question?}} \\
	      \textbf{Answer:} \clb{\texttt{Answer from LLM.}} \\
	      \clo{The final answer (\texttt{float/int/boolean}) is:} 
	};
	
\node[above left=0.25cm and -3.25cm of n3] (image1) {\includegraphics[height=0.7cm]{llm.jpg}};
\node[below= -0.25cm of image1] {\scriptsize LLM};
\node[above left=0.25cm and -3.25cm of n4] (image2) {\includegraphics[height=0.7cm]{llm.jpg}};
\node[below= -0.25cm of image2] {\scriptsize LLM};
\draw [arrow] (n1) edge[out=0, in=180, looseness=1.2] (n2);
\draw [arrow] (n3) edge[out=0, in=180, looseness=1.1] (n2);
\draw [arrow] (n1) -- (n3);
\draw [arrow] (n2) -- (n4);
	\node [fill=black, minimum width=0.05cm, minimum height=0.05cm, node distance=3mm, below left=0.25cm and 2.5cm of n4](c1) {};
	\node [right= -0.05cm of c1](cn1) {\scriptsize Signifier};
	\node [fill=orange, minimum width=0.05cm, minimum height=0.05cm, right= -0.01cm of cn1](c2) {};
	\node [right= -0.05cm of c2](cn2) {\scriptsize \clo{Memetic proxy}};
	\node [fill=blue, minimum width=0.05cm, minimum height=0.05cm, right= -0.01cm of cn2](c3) {};
	\node [right= -0.05cm of c3](cn3) {\scriptsize \clb{Input}};
\end{tikzpicture}
	\end{center}
\caption{ZS-STD prompt for FinQA}
\label{fig:standard_dual}
\end{figure*}

A standard template based prompting approach for question answering is a zero-shot standard dual (ZS-STD) prompt which has a LLM answering prompt and an answer extraction prompt. In the LLM answering prompt, the question is appended below the passage and then the trigger word ``Answer'' is added for LLM to generate the answer. The answer extraction prompt takes the LLM generated answer along with the question and append a memetic proxy phrase -- ``The final answer (\texttt{float/int/boolean}) is'' for extracting the final answer. The ZS-STD prompt for the FinQA dataset question answering is shown in Figure \ref{fig:standard_dual}. 

\subsection{Zero-shot Chain of Thoughts}
\begin{figure*}[htb]
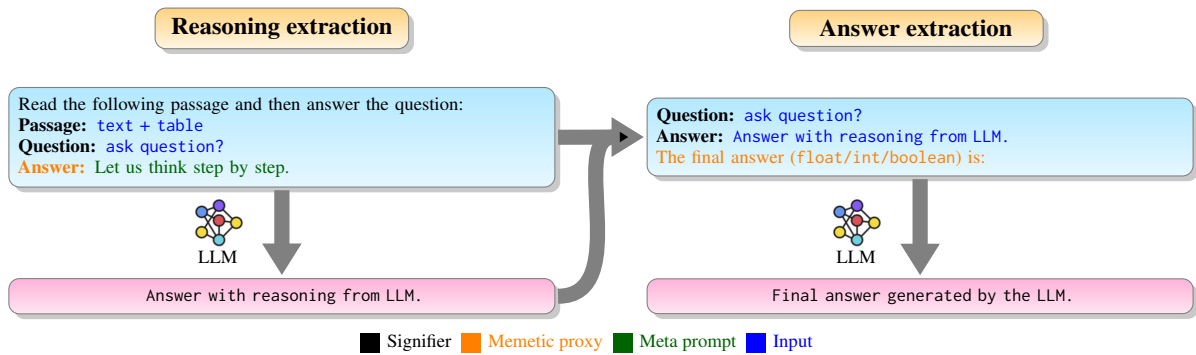

	\begin{center}
\begin{tikzpicture}[node distance=12mm]
	\node (a1) [heading] {\tn{Reasoning extraction}};
	\node (a2) [heading, right= 5cm of a1] {\tn{Answer extraction}};
	\node (n1) [prompt, node distance=5mm, below=of a1] {Read the following passage and then answer the question:\\ 
	\textbf{Passage:} \clb{\texttt{text} + \texttt{table}} \\
	\textbf{Question:} \clb{\texttt{ask question?}} \\
	\clo{\textbf{Answer:}} \cldg{Let us think step by step.}}; 
\node (n3) [response, below=of n1]
	      {\texttt{Answer with reasoning from LLM.}};
\node (n4) [response, right=of n3]
              {\texttt{Final answer generated by the LLM.}};
\node (n2) [prompt, node distance=13.5mm, above=of n4] 
	      {\textbf{Question:} \clb{\texttt{ask question?}} \\
	      \textbf{Answer:} \clb{\texttt{Answer with reasoning from LLM.}}\\
	      \clo{The final answer (\texttt{float/int/boolean}) is:} 
	};
	
\node[above left=0.25cm and -3.25cm of n3] (image1) {\includegraphics[height=0.7cm]{llm.jpg}};
\node[below= -0.25cm of image1] {\scriptsize LLM};
\node[above left=0.25cm and -3.25cm of n4] (image2) {\includegraphics[height=0.7cm]{llm.jpg}};
\node[below= -0.25cm of image2] {\scriptsize LLM};
\draw [arrow] (n1) edge[out=0, in=180, looseness=1.2] (n2);
\draw [arrow] (n3) edge[out=0, in=180, looseness=1.1] (n2);
\draw [arrow] (n1) -- (n3);
\draw [arrow] (n2) -- (n4);
	\node [fill=black, minimum width=0.05cm, minimum height=0.05cm, node distance=3mm, below left=0.25cm and 3.5cm of n4](c1) {};
	\node [right= -0.05cm of c1](cn1) {\scriptsize Signifier};
	\node [fill=orange, minimum width=0.05cm, minimum height=0.05cm, right= -0.01cm of cn1](c2) {};
	\node [right= -0.05cm of c2](cn2) {\scriptsize \clo{Memetic proxy}};
	\node [fill=darkgreen, minimum width=0.05cm, minimum height=0.05cm, right= -0.01cm of cn2](c4) {};
	\node [right= -0.05cm of c4](cn4) {\scriptsize \cldg{Meta prompt}};
	\node [fill=blue, minimum width=0.05cm, minimum height=0.05cm, right= -0.01cm of cn4](c5) {};
	\node [right= -0.05cm of c5](cn5) {\scriptsize \clb{Input}};
\end{tikzpicture}
	\end{center}
\caption{ZS-CoT prompt for FinQA}
\label{fig:zs_cot}
\end{figure*}

Similar to the zero-shot reasoners \cite{zs_cot}, zero-shot chain-of-thought (ZS-CoT) prompt is derived from the ZS-STD prompt by adding the reasoning trigger sentence --``Let us think step by step.'' after the word ``Answer:''. The answer extraction prompt of ZS-CoT is identical to the ZS-STD prompt. The ZS-CoT prompt for the FinQA dataset question answering is described in Figure \ref{fig:zs_cot}. 

All prompts for TATQA are identical to FinQA and for ConvFinQA dataset, the prompts are slightly modified to handle conversational questions as shown in Appendix \ref{app:conv-prompt}. 

\section{Experiments}
\subsection{Experimental Design}
\textbf{Datasets:}
We conduct our experiments on three financial question answering datasets FinQA \cite{finqa}, ConvFinQA \cite{convfinqa} and TATQA \cite{tatqa} as summarized in Table \ref{tab:datasets}. For our evaluations, we use the test split of FinQA, while for ConvFinQA and TATQA we use the dev set as answers for test splits of these datasets are not available. The evaluations for TATQA are restricted to questions of \textit{arithmetic} type. The question answering task is to answer the questions using the passage containing text and table content. The table content is represented in a textual format using the strategy adopted in \cite{chen}. In the textual format, the table columns are separated by `|', the rows are separated by `\textbackslash n' and the empty cell are filled with `-'.     

\textbf{Large Language Models:} 
We experimented with three Azure OpenAI\footnote{https://oai.azure.com/} LLMs text-davinci-003, gpt-3.5-turbo, gpt-4. The Python programs generated using LLMs are executed using Python function \mfun{exec}. The domain specific programs are executed using the Python script provided by FinQA.\footnote{https://github.com/czyssrs/FinQA} In order to achieve a more precise and predictable outcome, the LLM parameters are set as follows: \textit{temperature} = 0, \textit{top\_prob} = 0.95, \textit{max\_tokens} = 1000.     

\textbf{Evaluation Metrics:} 
For all the financial datasets -- FinQA, ConvFinQA and TATQA, we implement the evaluation strategy discussed in program of thoughts prompting \cite{pot} on Github\footnote{https://github.com/wenhuchen/Program-of-Thoughts} with slight modifications. The LLM responses are varying in nature for questions with answers in \textit{thousands, millions,} and \textit{percentage}. Examples: for the gold answer 7 million, the gpt response may be 7 million or 7,000,000; for the gold answer 23\%, the gpt response may be 23\% or 0.23. The evaluation strategy is modified to handle such cases. We relax the evaluations for ZS-CoT \cite{zs_cot} and standard dual prompting because LLMs using these prompting techniques generate answers instead of programs. Since LLMs cannot perform precise mathematical calculations (especially with high-precision floats and large numbers),we provide a tolerance while comparing the gpt final answer with the gold answer. The evaluation is implemented using the Python function \mfun{isclose} with a relative tolerance (rel\_tol) of 0.001. The \mfun{isclose} functionality returns \mfun{True} while comparing the gpt final answer $(\hat{a})$ with the gold answer $(\tilde{a})$ if and only if the condition     
\[
\abs(\hat{a} - \tilde{a}) \leq \text{rel\_tol} * \max(\abs(\hat{a}), \abs(\tilde{a}))
\]
is satisfied.

\textbf{Baselines:} 
We consider two baselines for zero-shot prompting setting: ZS-STD prompt and ZS-CoT prompt. These zero-shot prompting techniques are evaluated with all three Azure OpenAI models (text-davinci-003, gpt-3.5-turbo, gpt-4) on all three financial datasets (FinQA, ConvFinQA and TATQA).   

\begin{table*}[htb]
\centering
\begin{threeparttable}
\begin{tabular}{lllll}
\hline
	\textbf{Dataset} & \textbf{Split} & \textbf{Example} & \textbf{Input} & \textbf{Output}\\
\hline
	FinQA & Test & 1147 &  Table + Text + Question & Number+Binary \\
	ConvFinQA & Dev & 421 &  Table + Text + Multi-turn Question & Number+Binary \\
	TATQA & Dev \tnote{$\dag$} & 718 &  Table + Text + Question & Number+Binary \\
\hline
\end{tabular}
\begin{tablenotes}
      \item[$\dag$] {\footnotesize Only arithmetic questions from the Dev split of TATQA.}
    \end{tablenotes}
\end{threeparttable}
\caption{\label{tab:datasets}Financial question answering datasets for evaluation}
\end{table*}

\begin{table*}[hbt]
\centering
\begin{tabular}{lccc}
\hline
	\textbf{Models} & \textbf{FinQA} & \textbf{ConvFinQA} & \textbf{TATQA}\\
\hline
	ZS-STD (text-davinci-003) & 22.58 & 13.30 & 39.97 \\
	ZS-CoT (text-davinci-003) & 41.15 & 27.08 & 68.94 \\
	ZS-FinDSL (text-davinci-003) & 56.76 & 52.02 & 68.25 \\
	ZS-FinPYT (text-davinci-003) & \textbf{66.60} & \textbf{60.30} & \textbf{78.40} \\
\hline
	ZS-STD (gpt-3.5-turbo) & 32.26 & 47.74 & 49.03 \\
	ZS-CoT (gpt-3.5-turbo) & 53.01 & 52.49 & 74.09 \\
	ZS-FinDSL (gpt-3.5-turbo) & 61.12 & 60.81 & 77.86 \\
	ZS-FinPYT (gpt-3.5-turbo) & \textbf{66.52} & \textbf{67.45} & \textbf{85.00}\\
\hline
	ZS-STD (gpt-4) & 63.64 & 72.45 & 77.58 \\
	ZS-CoT (gpt-4) & 74.19 & 75.30  & 90.11 \\
	ZS-FinDSL (gpt-4) & 77.33 & \textbf{77.67} & 90.53 \\
	ZS-FinPYT (gpt-4) & \textbf{77.51} & 76.95 & \textbf{93.00} \\
\hline
\end{tabular}
\caption{\label{tab:fin-accuracy}
	Comparison results of various models on different datasets.
}
\end{table*}

\begin{table*}[hbt]
\centering
\begin{threeparttable}
\begin{tabular}{lccc}
\hline
	\textbf{Models} & \textbf{FinQA} & \textbf{ConvFinQA} & \textbf{TATQA}\\
\hline
	Few-shot PoT (text-davinci-003)\tnote{$\star$} & 72.27 & 69.35 & 83.21 \\
	ZS-FinPYT (text-davinci-003) & 66.60 & 60.30 & 78.40\\
\hline
	Few-shot PoT (gpt-3.5-turbo)\tnote{$\star$} & 67.39 & 65.79 & 74.75 \\
	ZS-FinPYT (gpt-3.5-turbo) & 66.52 & 67.45 & 85.00\\
\hline
	Few-shot PoT (gpt-4)\tnote{$\star$} & 78.46 & 82.42 & 91.89 \\
	ZS-FinPYT (gpt-4) & 77.51 & 76.95 & 93.00 \\
\hline
\end{tabular}
\begin{tablenotes}
      \item[$\star$] {\footnotesize Few-shot PoT uses 4-shots selected from the few-shots used in \cite{pot}.}
    \end{tablenotes}
\end{threeparttable}
\caption{\label{tab:fin-pot-accuracy}
	Performance of ZS-FinPYT and few-shot PoT on different datasets.
}
\end{table*}

\subsection{Main Results}
The evaluation results for the proposed prompting techniques ZS-FinPYT and ZS-FinDSL along with the baselines ZS-STD prompt and ZS-CoT are summarized in Table \ref{tab:fin-accuracy}. The ZS-FinPYT and ZS-FinDSL methods significantly outperform the ZS-STD prompt for all datasets and across all LLMs. The ZS-FinPYT achieves 4.5\% to 47\% and the ZS-FinDSL achieves 5.22\% to 38.72\% improvement in accuracy over ZS-STD. The increase in accuracy for text-davinci and gpt-3.5 are are much higher than that for gpt-4 as for gpt-4 the base model performs reasonably well. These results indicate that our prompts are able to induce the required reasoning and successfully output the required Python programs or domain specific languages. 

Both methods also made significant improvements over the ZS-CoT method for text-davinci-003 and gpt-3.5-turbo, with the ZS-FinPYT achieving 3\% to 33.22\% and the ZS-FinDSL achieving 0\% to 24.94\% improvement over the ZS-CoT on different datasets. For gpt-4, our approach slightly outperforms the ZS-CoT for all datasets with improvements in the range of 1.5-3.5\%. However, it is important to highlight that ZS-CoT lacks the ability to provide precise answers, and its accuracy is measured using a relaxed metric, while our method generates precise answers and an exact metric is used to measure accuracy.

In general, the ZS-FinPYT approach gave better results than ZS-FinDSL for the text-davinci-and gpt-3.5-turbo models for the different datasets. For gpt-4 both methods are comparable.  

We also carried out an evaluation of OpenAI models using few-shot PoT prompting, as shown in Table \ref{tab:fin-pot-accuracy}. The comparisons indicate the excellent performance of our zero-shot method as we are within 10\% of the few-shot and in many cases almost the same and for few cases even surpassing the few-shot performance.   

\begin{table*}[htb]
\centering
\begin{tabular}{lcccc}
\hline
	\textbf{Methods} & \multicolumn{2}{c}{\textbf{ZS-FinPYT}} & \multicolumn{2}{c}{\textbf{ZS-FinDSL}}\\
\hline
& \phantom{q} \textbf{gpt-4} \phantom{q}& \phantom{q} \textbf{gpt-3.5-turbo} \phantom{q}  & \phantom{q} \textbf{gpt-4} \phantom{q} & \phantom{q} \textbf{gpt-3.5-turbo} \\
\hline
overall accuracy & 77.51 & 66.52 & 77.33 & 61.12\\
\hline
\multicolumn{5}{l}{\textbf{Performance on table and text}}\\
\hline
table-only questions & 80.91 & 71.36 & 81.36 & 63.94 \\
text-only questions & 74.45 & 58.39 & 73.36 & 60.22 \\
table-text questions & 67.44 & 55.81 & 68.22 & 48.84 \\
\hline
\multicolumn{5}{l}{\textbf{Performance regarding program steps}}\\
\hline
1 step programs & 80.73 & 69.27 & 79.82 & 62.08 \\
2 step programs & 77.02 & 64.79 & 77.26 & 63.08 \\
>2 step programs & 54.76 & 53.57 & 58.33 & 44.05 \\
\hline
\multicolumn{5}{l}{\textbf{Performance regarding question types}}\\
\hline
boolean questions & 90.00& 95.00 & 85.00 & 45.00 \\
numerical questions & 77.28 & 66.02 & 77.20 & 61.40 \\
\hline
\end{tabular}
\caption{\label{tab:perf-accuracy}
Performance breakdown of various models on FinQA dataset.
}
\end{table*}

\subsection{Performance Analysis}
We conduct a performance analysis on FinQA dataset for two models gpt-4, gpt-3.5-turbo, see Table \ref{tab:perf-accuracy} for details. The FinQA questions are divided into various categories to gain further insights.

\textbf{Performance on text and table questions:}The FinQA questions are divided into three sets depending on where the information required to answer the question is available: table only questions, text-only questions, text-table questions. 

\textbf{Performance regarding program steps:} The FinQA questions are divided into three sets based on number of steps required to provide the answer: 1 step program, 2 step program and >2 step program. 

\textbf{Performance regarding question types:} The FinQA questions are divided into numerical and boolean type questions.  

The key findings are listed below:

\textbf{The models achieve the highest accuracy on table-only questions.} As tables are structured and the tables in this dataset are simple, it maybe easier for the LLMs to more accurately extract the values as compared to extracting from unstructured text. 
 
\textbf{Question with multi-hop reasoning are challenging.}  As would be expected both models find it easier to answer questions with one or two hop-reasoning as compared to questions needing more than two hop reasoning. 

\textbf{Numerical questions are more challenging as compared to boolean questions.} In general, gpt-4 and gpt-3.5-turbo models excel in answering boolean questions over arithmetic questions. However, gpt-3.5-turbo's performance declines with ZS-FinDSL prompt  for boolean questions as compared to arithmetic questions. Examination of a few cases indicated that gpt-3.5-turbo has greater difficulty in writing DSL programs correctly for boolean questions.  

\subsection{Error Analysis}
We sampled 50 test cases from FinQA dataset results of text-davinci-003 model and examined in detail the entire output of the system to get further insight into the obtained results. As expected, ZS-STD prompt results in brief answers with a sentence or value as the output without providing any details on the reasoning potentially contributing to its poor performance. On the other hand, LLM responses with ZS-CoT details out the reasoning behind the answers and shows significantly better performance than ZS-STD. However, arithmetic errors results into a substantial drop in performance for both ZS-STD prompt and ZS-CoT.       

The ZS-FinPYT and ZS-FinDSL approaches demonstrated detailed reasoning. In the case of ZS-FinPYT the task of writing a Python program triggers reasoning while in the case of ZS-FinDSL there are two prompts where the first prompt is a meta prompt that drives the reasoning similar to ZS-CoT. These techniques produce programs instead of answers for questions and therefore, mitigate arithmetic errors. Hence, these proposed techniques significantly outperforms ZS-CoT. The ZS-FinDSL performance is lower than ZS-FinPYT because the program extraction step fails for some cases where the reasoning step is correct. One possible explanation could be that the GPT systems have likely been trained on huge amounts of Python programs and hence can generate Python program efficiently where as for ZS-FinDSL the instruction contains the information on how to write out the domain specific program. This may be driving the slightly higher error rate of the ZS-FinDSL. Some demonstrative examples supporting these observations may be found in Appendix \ref{app:error-analysis}.    

\section{Conclusion}
We proposed zero-shot prompting techniques to answer complex questions requiring multi-hop numerical reasoning over financial reports. The prompts guide the LLM to encode the required reasoning into a program that is executed by a program interpreter. The approach demonstrated excellent results on three financial datasets, achieving significant improvement over the respective baselines.We hope that our work will motivate a principled approach to prompt design with other LLMs. 

\section*{Limitations}
In this paper, we only experiment with the GPT-x series of LLMs. While this work shows the tremendous potential for zero-shot financial reasoning with LLMs, it is possible that better performance may be obtained with other LLMs. Moreover, the prompts we have proposed are designed to address specific problems observed with the three GPT models considered in this work. Other LLMs may behave differently and will likely need modification to the prompts to work effectively. 

While we experiment and find zero-shot prompts that are effective for both ZS-FinPYT and ZS-FinDSL, and the error analysis provided insights into failures, there are also unexplained failures in reasoning and more research is needed to understand the behavior  of LLMs for certain cases. For ZS-FinDSL, we observed some patterns that result in failure of program extraction. However, it is unclear what drives these failures and we leave that for future work.

For cases where the reasoning was incorrect, the system may provide an explanation with a high-level of confidence. Our prompt currently does not address or control for such behavior. In practice, this can pose challenges for practical use in real world systems.

\section*{Disclaimer}
The views reflected in this article are the views of the authors and do not necessarily reflect the views of the global EY organization or its member firms.

\bibliography{references}
\bibliographystyle{acl_natbib}

\appendix
\section{Prompts for ConvFinQA}
\label{app:conv-prompt}
The ConvFinQA prompts are slightly modified to handle conversational questions. 

\begin{figure*}[htb]
	\begin{center}
\begin{tikzpicture}[node distance=12mm]
	\node (n1) [prompt] {Read the following text and table, and then answer the last question by writing a Python code:\\ 
	\textbf{Passage:} \clb{\texttt{text} + \texttt{table}} \\
	\textbf{Questions:} \clb{\texttt{ask a series of questions?}} \\
	\textbf{Last Question:} \clb{\texttt{ask last question of the series?}} \\
	Answer the last question by following the below instructions. \\
	\textbf{Instructions:} \\  
		\phantom{new} \clm{Define the Python variable which must begin with a character. \\}
		\phantom{new} Assign values to variables required for the calculation. \\
		\phantom{new} Create Python variable "ans" and assign the final answer \\
		\phantom{new} (bool/float) to the variable "ans".\\
		\phantom{new} \clm{Don't include non-executable statements and include them as \\
		\phantom{new} part of comments.} \clo{\#Comment: $\ldots$}\\ 
	Python executable code is: \\
	\clo{\textbf{\#Python}}
}; 

\node (n3) [response, below=of n1]
	      {\texttt{Python code from the LLM.}};
\draw [arrow] (n1) --  (n3);
\node[above left=0.25cm and -3.25cm of n3] (image) {\includegraphics[height=0.7cm]{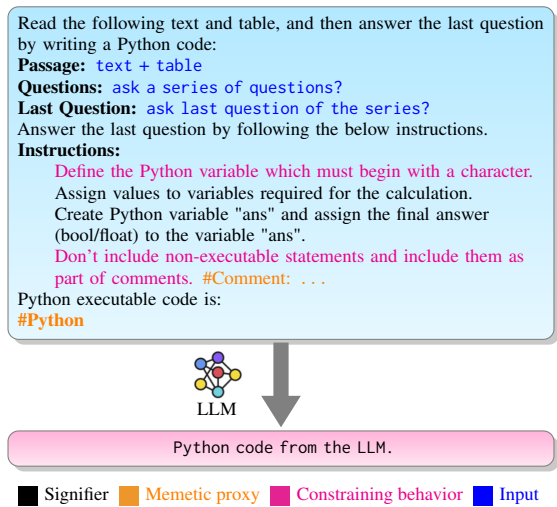}};
\node[below= -0.25cm of image] {\scriptsize LLM};
	\node [fill=black, minimum width=0.05cm, minimum height=0.05cm, node distance=3mm, below left=0.25cm and -0.4cm of n3](c1) {};
	\node [right= -0.05cm of c1](cn1) {\scriptsize Signifier};
	\node [fill=cyan!05!orange, minimum width=0.05cm, minimum height=0.05cm, right= -0.01cm of cn1](c2) {};
	\node [right= -0.05cm of c2](cn2) {\scriptsize \clo{Memetic proxy}};
	\node [fill=cyan!05!magenta, minimum width=0.05cm, minimum height=0.05cm, right= -0.01cm of cn2](c3) {};
	\node [right= -0.05cm of c3](cn3) {\scriptsize \clm{Constraining behavior}};
	\node [fill=blue, minimum width=0.05cm, minimum height=0.05cm, right= -0.01cm of cn3](c4) {};
	\node [right= -0.05cm of c4](cn4) {\scriptsize \clb{Input}};
\end{tikzpicture}
	\end{center}
\caption{ZS-FinPYT (gpt-4) prompt for ConvFinQA}
\label{fig:gpt_zs_fin_pyt}
\end{figure*}

\begin{figure*}[htb]
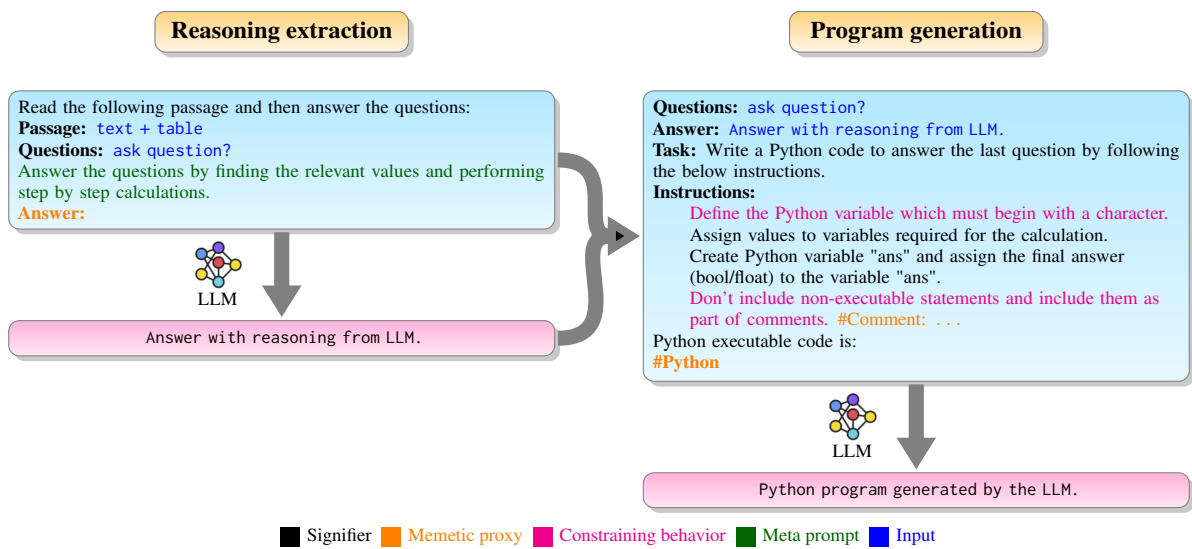

	\begin{center}
\begin{tikzpicture}[node distance=12mm]
	\node (a1) [heading] {\tn{Reasoning extraction}};
	\node (a2) [heading, right= 5cm of a1] {\tn{Program generation}};
	\node (n1) [prompt, node distance=5mm, below=of a1] {Read the following passage and then answer the questions:\\ 
	\textbf{Passage:} \clb{\texttt{text} + \texttt{table}} \\
	\textbf{Questions:} \clb{\texttt{ask question?}} \\
	\cldg{Answer the questions by finding the relevant values and performing step by step calculations.\\}
	\clo{\textbf{Answer:} }}; 
\node (n2) [prompt, node distance=5mm, below=of a2] 
	      {\textbf{Questions:} \clb{\texttt{ask question?}} \\
	      \textbf{Answer:} \clb{\texttt{Answer with reasoning from LLM.}}\\
	      \textbf{Task:} Write a Python code to answer the last question by following the below instructions.\\
	\textbf{Instructions:} \\  
		\phantom{new} \clm{Define the Python variable which must begin with a character. \\}
		\phantom{new} Assign values to variables required for the calculation. \\
		\phantom{new} Create Python variable "ans" and assign the final answer \\
		\phantom{new} (bool/float) to the variable "ans".\\
		\phantom{new} \clm{Don't include non-executable statements and include them as \\
		\phantom{new} part of comments.} \clo{\#Comment: $\ldots$}\\ 
	Python executable code is: \\
	\clo{\textbf{\#Python}}
};
\node (n4) [response, below=of n2]
              {\texttt{Python program generated by the LLM.}};
\node (n3) [response, below=of n1]
	      {\texttt{Answer with reasoning from LLM.}};
\node[above left=0.25cm and -3.25cm of n3] (image1) {\includegraphics[height=0.7cm]{llm.jpg}};
\node[below= -0.25cm of image1] {\scriptsize LLM};
\node[above left=0.25cm and -3.25cm of n4] (image2) {\includegraphics[height=0.7cm]{llm.jpg}};
\node[below= -0.25cm of image2] {\scriptsize LLM};
\draw [arrow] (n1) edge[out=0, in=180, looseness=1.5] (n2);
\draw [arrow] (n3) edge[out=0, in=180, looseness=1.5] (n2);
\draw [arrow] (n1) -- (n3);
\draw [arrow] (n2) -- (n4);
	\node [fill=black, minimum width=0.05cm, minimum height=0.05cm, node distance=3mm, below left=0.25cm and 4.5cm of n4](c1) {};
	\node [right= -0.05cm of c1](cn1) {\scriptsize Signifier};
	\node [fill=orange, minimum width=0.05cm, minimum height=0.05cm, right= -0.01cm of cn1](c2) {};
	\node [right= -0.05cm of c2](cn2) {\scriptsize \clo{Memetic proxy}};
	\node [fill=magenta, minimum width=0.05cm, minimum height=0.05cm, right= -0.01cm of cn2](c3) {};
	\node [right= -0.05cm of c3](cn3) {\scriptsize \clm{Constraining behavior}};
	\node [fill=darkgreen, minimum width=0.05cm, minimum height=0.05cm, right= -0.01cm of cn3](c4) {};
	\node [right= -0.05cm of c4](cn4) {\scriptsize \cldg{Meta prompt}};
	\node [fill=blue, minimum width=0.05cm, minimum height=0.05cm, right= -0.01cm of cn4](c5) {};
	\node [right= -0.05cm of c5](cn5) {\scriptsize \clb{Input}};
\end{tikzpicture}
	\end{center}
\caption{ZS-FinPYT (gpt-3.5-turbo, text-davinci-003) prompt for ConvFinQA}
\label{fig:czs_fin_pyt}
\end{figure*}

\textbf{ZS-FinPYT for ConvFinQA:}
For gpt-4, we use a single prompt where the last question in the series of questions is clearly marked and the system is instructed to answer the last questions as shown in Figure \ref{fig:gpt_zs_fin_pyt}. For gpt-3.5-turbo and text-davinci-003, we use dual prompt approach which consists of a reasoning extraction prompt and a program generation prompt, see Figure \ref{fig:czs_fin_pyt}. The reasoning extraction prompt is there to generate answers with reasoning for all the questions in a conversation, and the program generation prompt generates a Python program answering the last question.  

\begin{figure*}[htb]
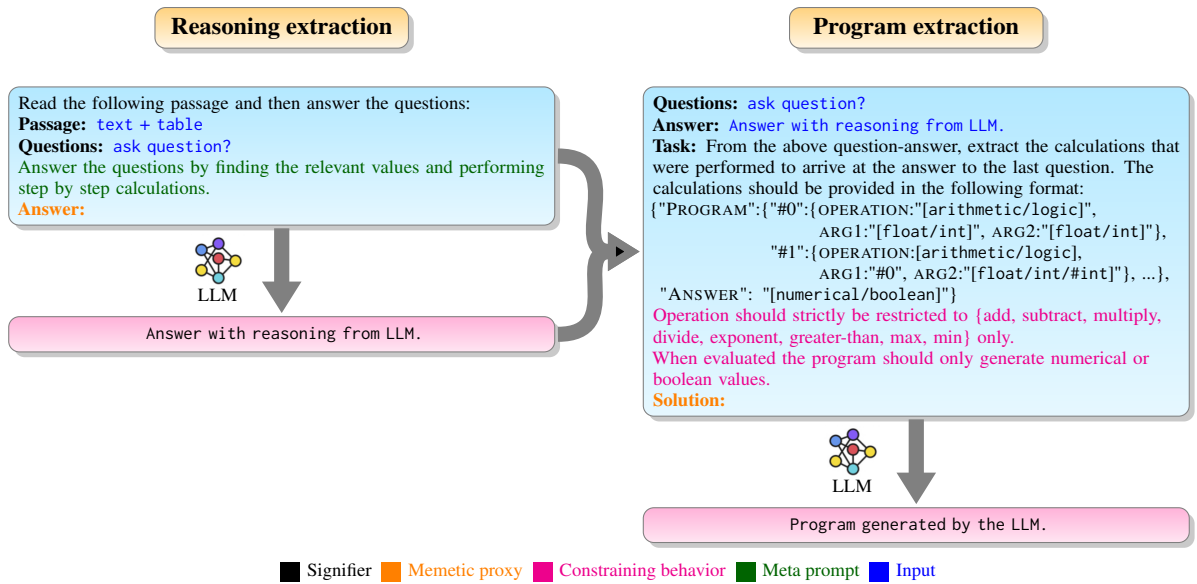

	\begin{center}
\begin{tikzpicture}[node distance=12mm]
	\node (a1) [heading] {\tn{Reasoning extraction}};
	\node (a2) [heading, right= 5cm of a1] {\tn{Program extraction}};
	\node (n1) [prompt, node distance=5mm, below=of a1] {Read the following passage and then answer the questions:\\ 
	\textbf{Passage:} \clb{\texttt{text} + \texttt{table}} \\
	\textbf{Questions:} \clb{\texttt{ask question?}} \\
	\cldg{Answer the questions by finding the relevant values and performing step by step calculations.\\}
	\clo{\textbf{Answer:} }}; 
\node (n2) [prompt, node distance=5mm, below=of a2] 
	      {\textbf{Questions:} \clb{\texttt{ask question?}} \\
	      \textbf{Answer:} \clb{\texttt{Answer with reasoning from LLM.}}\\
	      \textbf{Task:} From the above question-answer, extract the calculations that were performed to arrive at the answer to the last question. The calculations should be provided in the following format:\\
	      \{"\textsc{Program}":\{"\#0":\{\textsc{operation}:"[\texttt{arithmetic/logic}]",\\
	      \phantom{\{"\textsc{Program}":\{"\#0":} \textsc{arg1}:"[\texttt{float/int}]", \textsc{arg2}:"[\texttt{float/int}]"\}, \\
	      \phantom{\{"\textsc{Program}":} "\#1":\{\textsc{operation}:[\texttt{arithmetic/logic}], \\
	      \phantom{\{"\textsc{Program}":\{"\#0":} \textsc{arg1}:"\#0", \textsc{arg2}:"[\texttt{float/int/\#int}]"\}, ...\}, \\
	      \textsc{ "Answer"}: "[\texttt{numerical/boolean}]"\} \\
	      \clm{Operation should strictly be restricted to \{add, subtract, multiply, divide, exponent, greater-than, max, min\} only. \\
	      When evaluated the program should only generate numerical or boolean values.\\}
	\clo{\textbf{Solution:}} };
	
\node (n4) [response, below=of n2]
              {\texttt{Program generated by the LLM.}};
\node (n3) [response, below=of n1]
	      {\texttt{Answer with reasoning from LLM.}};
\node[above left=0.25cm and -3.25cm of n3] (image1) {\includegraphics[height=0.7cm]{llm.jpg}};
\node[below= -0.25cm of image1] {\scriptsize LLM};
\node[above left=0.25cm and -3.25cm of n4] (image2) {\includegraphics[height=0.7cm]{llm.jpg}};
\node[below= -0.25cm of image2] {\scriptsize LLM};
\draw [arrow] (n1) edge[out=0, in=180, looseness=1.5] (n2);
\draw [arrow] (n3) edge[out=0, in=180, looseness=1.5] (n2);
\draw [arrow] (n1) -- (n3);
\draw [arrow] (n2) -- (n4);
	\node [fill=black, minimum width=0.05cm, minimum height=0.05cm, node distance=3mm, below left=0.25cm and 4.5cm of n4](c1) {};
	\node [right= -0.05cm of c1](cn1) {\scriptsize Signifier};
	\node [fill=orange, minimum width=0.05cm, minimum height=0.05cm, right= -0.01cm of cn1](c2) {};
	\node [right= -0.05cm of c2](cn2) {\scriptsize \clo{Memetic proxy}};
	\node [fill=magenta, minimum width=0.05cm, minimum height=0.05cm, right= -0.01cm of cn2](c3) {};
	\node [right= -0.05cm of c3](cn3) {\scriptsize \clm{Constraining behavior}};
	\node [fill=darkgreen, minimum width=0.05cm, minimum height=0.05cm, right= -0.01cm of cn3](c4) {};
	\node [right= -0.05cm of c4](cn4) {\scriptsize \cldg{Meta prompt}};
	\node [fill=blue, minimum width=0.05cm, minimum height=0.05cm, right= -0.01cm of cn4](c5) {};
	\node [right= -0.05cm of c5](cn5) {\scriptsize \clb{Input}};
\end{tikzpicture}
	\end{center}
\caption{ZS-FinDSL prompt for ConvFinQA}
\label{fig:czs_fin_dsl}
\end{figure*}

\textbf{ZS-FinDSL for ConvFinQA:} The ZS-FinDSL for ConvFinQA , see Figure \ref{fig:czs_fin_dsl}, is a dual prompt which consists of a reasoning prompt and a program extraction prompt that are similar to the corresponding prompts for FinQA. The reasoning prompt instructs the LLM to generate answers with reasoning for all questions in a conversation. The program extraction prompt is instructing the LLM to generate program for performing calculations to answer the last question.  

\begin{figure*}[htb]
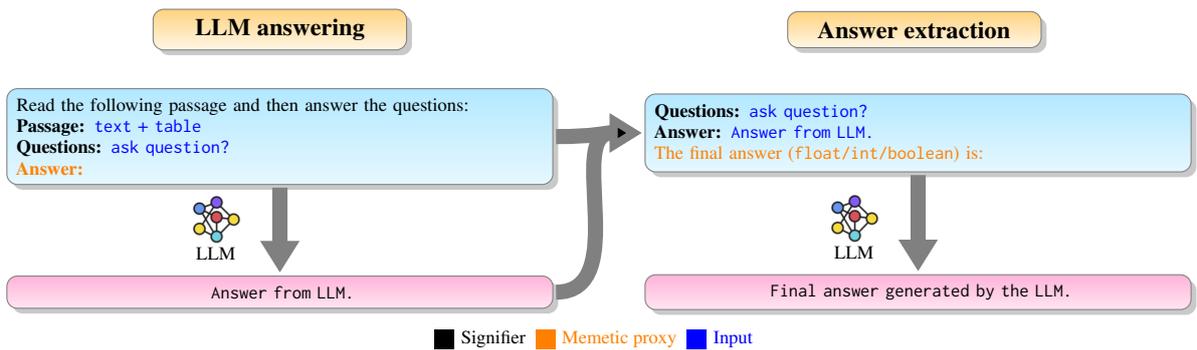

	\begin{center}
\begin{tikzpicture}[node distance=12mm]
	\node (a1) [heading] {\tn{LLM answering}};
	\node (a2) [heading, right= 5cm of a1] {\tn{Answer extraction}};
	\node (n1) [prompt, node distance=5mm, below=of a1] {Read the following passage and then answer the questions:\\ 
	\textbf{Passage:} \clb{\texttt{text} + \texttt{table}} \\
	\textbf{Questions:} \clb{\texttt{ask question?}} \\
	\clo{\textbf{Answer:}} }; 
\node (n3) [response, below=of n1]
	      {\texttt{Answer from LLM.}};
\node (n4) [response, right=of n3]
              {\texttt{Final answer generated by the LLM.}};
\node (n2) [prompt, node distance=13.5mm, above=of n4] 
	      {\textbf{Questions:} \clb{\texttt{ask question?}} \\
	      \textbf{Answer:} \clb{\texttt{Answer from LLM.}} \\
	      \clo{The final answer (\texttt{float/int/boolean}) is:} 
	};
	
\node[above left=0.25cm and -3.25cm of n3] (image1) {\includegraphics[height=0.7cm]{llm.jpg}};
\node[below= -0.25cm of image1] {\scriptsize LLM};
\node[above left=0.25cm and -3.25cm of n4] (image2) {\includegraphics[height=0.7cm]{llm.jpg}};
\node[below= -0.25cm of image2] {\scriptsize LLM};
\draw [arrow] (n1) edge[out=0, in=180, looseness=1.2] (n2);
\draw [arrow] (n3) edge[out=0, in=180, looseness=1.1] (n2);
\draw [arrow] (n1) -- (n3);
\draw [arrow] (n2) -- (n4);
	\node [fill=black, minimum width=0.05cm, minimum height=0.05cm, node distance=3mm, below left=0.25cm and 2.5cm of n4](c1) {};
	\node [right= -0.05cm of c1](cn1) {\scriptsize Signifier};
	\node [fill=orange, minimum width=0.05cm, minimum height=0.05cm, right= -0.01cm of cn1](c2) {};
	\node [right= -0.05cm of c2](cn2) {\scriptsize \clo{Memetic proxy}};
	\node [fill=blue, minimum width=0.05cm, minimum height=0.05cm, right= -0.01cm of cn2](c3) {};
	\node [right= -0.05cm of c3](cn3) {\scriptsize \clb{Input}};
\end{tikzpicture}
	\end{center}
\caption{ZS-STD prompt for ConvFinQA}
\label{fig:czs_std}
\end{figure*}

\begin{figure*}[htb]
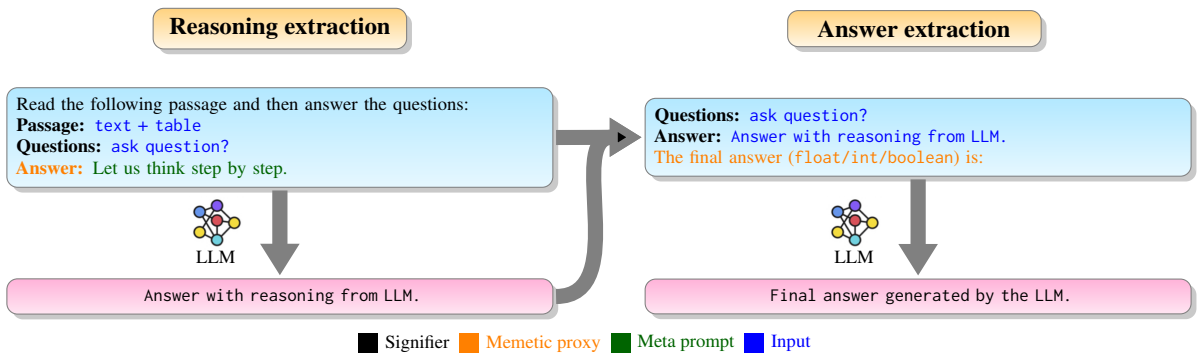

	\begin{center}
\begin{tikzpicture}[node distance=12mm]
	\node (a1) [heading] {\tn{Reasoning extraction}};
	\node (a2) [heading, right= 5cm of a1] {\tn{Answer extraction}};
	\node (n1) [prompt, node distance=5mm, below=of a1] {Read the following passage and then answer the questions:\\ 
	\textbf{Passage:} \clb{\texttt{text} + \texttt{table}} \\
	\textbf{Questions:} \clb{\texttt{ask question?}} \\
	\clo{\textbf{Answer:}} \cldg{Let us think step by step.}}; 
\node (n3) [response, below=of n1]
	      {\texttt{Answer with reasoning from LLM.}};
\node (n4) [response, right=of n3]
              {\texttt{Final answer generated by the LLM.}};
\node (n2) [prompt, node distance=13.5mm, above=of n4] 
	      {\textbf{Questions:} \clb{\texttt{ask question?}} \\
	      \textbf{Answer:} \clb{\texttt{Answer with reasoning from LLM.}}\\
	      \clo{The final answer (\texttt{float/int/boolean}) is:} 
	};
	
\node[above left=0.25cm and -3.25cm of n3] (image1) {\includegraphics[height=0.7cm]{llm.jpg}};
\node[below= -0.25cm of image1] {\scriptsize LLM};
\node[above left=0.25cm and -3.25cm of n4] (image2) {\includegraphics[height=0.7cm]{llm.jpg}};
\node[below= -0.25cm of image2] {\scriptsize LLM};
\draw [arrow] (n1) edge[out=0, in=180, looseness=1.2] (n2);
\draw [arrow] (n3) edge[out=0, in=180, looseness=1.1] (n2);
\draw [arrow] (n1) -- (n3);
\draw [arrow] (n2) -- (n4);
	\node [fill=black, minimum width=0.05cm, minimum height=0.05cm, node distance=3mm, below left=0.25cm and 3.5cm of n4](c1) {};
	\node [right= -0.05cm of c1](cn1) {\scriptsize Signifier};
	\node [fill=orange, minimum width=0.05cm, minimum height=0.05cm, right= -0.01cm of cn1](c2) {};
	\node [right= -0.05cm of c2](cn2) {\scriptsize \clo{Memetic proxy}};
	\node [fill=darkgreen, minimum width=0.05cm, minimum height=0.05cm, right= -0.01cm of cn2](c4) {};
	\node [right= -0.05cm of c4](cn4) {\scriptsize \cldg{Meta prompt}};
	\node [fill=blue, minimum width=0.05cm, minimum height=0.05cm, right= -0.01cm of cn4](c5) {};
	\node [right= -0.05cm of c5](cn5) {\scriptsize \clb{Input}};
\end{tikzpicture}
	\end{center}
\caption{ZS-CoT prompt for ConvFinQA}
\label{fig:czs_cot}
\end{figure*}

\textbf{ZS-STD and ZS-CoT for ConvFinQA:} 
The LLM answering prompt of ZS-STD, see Figure \ref{fig:czs_std}, and the reasoning extraction prompt of ZS-CoT, see Figure \ref{fig:czs_cot}, are instructing the LLM to answer the questions of a conversation. Then the answer extraction prompt of both of these technique extract the final answer.  

\section{Error Analysis Examples}
\label{app:error-analysis}
We show some examples from FinQA dataset with the corresponding responses from the text-davinci-003 model under various prompts. These examples demonstrate successful attempts and failure cases under various prompts. 

We begin with showing some examples where ZS-FinDSL (text-davinci-003) system generates correct reasoning and the corresponding program generation succeeded, see Figure \ref{fig:fin_dsl_1} and Figure \ref{fig:fin_dsl_2}. Similarly, Figure \ref{fig:fin_pyt_1} and Figure \ref{fig:fin_pyt_2} show successful Python program generation by the system ZS-FinPYT (text-davinci-003). 
\begin{figure*}[htb]
\begin{center}
\begin{tikzpicture}[node distance=12mm]
	\node (n1) [example_prompt] {
\qaid{UPS/2010/page\_52.pdf-1}

\qabox{gray}{
\textbf{Passage:} contractual commitments we have contractual obligations and commitments in the form of capital leases , operating leases $\ldots$ \\
Commitment Type | 2011 | 2012 | 2013 | 2014 | 2015 | After 2016 | Total\\
Capital Leases | \$18 | \$19 | \$19 | \$20 | \$21 | \$112 | \$209\\
Other Liabilities | 69 | 67 | 64 | 58 | 43 | 38 | 339\\
Total | \$2,944 | \$1,334 | \$3,515 | \$2,059 | \$820 | \$12,884 | \$23,556\\
\textbf{Question:} what percentage of total expected cash outflow to satisfy contractual obligations and commitments as of december 31 , 2010 are due in 2012?  
}

\gptbox{blue}{
\textbf{Gold Program:} divide(1334, 23556) \\ 
\textbf{Gold Answer:} 0.05663 
}

\gptbox{violet}{
\textbf{ZS-FinDSL Reasoning Extraction Prompt Response:}\\
Total expected cash outflow to satisfy contractual obligations and commitments as of December 31, 2010 = \$23,556 (million), Cash outflow due in 2012 = \$1,334 (million), Percentage of total expected cash outflow due in 2012 = (1,334/23,556) x 100 = 5.65\% \\
\textbf{ZS-FinDSL Program Extraction Prompt Response:}\\
\{"Program": \{"\#0":\{operation:"divide", arg1:"1,334", arg2:"23,556"\}, "Answer": "282"\} \\
\textbf{ZS-FinDSL Program:} divide(1334, 23556)\\
\textbf{ZS-FinDSL Executed Answer:} 0.05663 
}
}; 
\end{tikzpicture}
\end{center}
\caption{An example where ZS-FinDSL extracted correct reasoning and successfully generated the DSL program.\label{fig:fin_dsl_1}}
\end{figure*}

\begin{figure*}[htb]
\begin{center}
\begin{tikzpicture}[node distance=12mm]
	\node (n1) [example_prompt] {
\qaid{HUM/2009/page\_105.pdf-2}

\qabox{gray}{
\textbf{Passage:} humana inc . notes to consolidated financial statements 2014 ( continued ) value , or the excess of the market value $\ldots$ \\
-  | 2009 | 2008 | 2007\\
Weighted-average fair value at grant date | \$14.24 | \$17.95 | \$21.07\\
Expected option life (years) | 4.6 | 5.1 | 4.8\\
Expected volatility | 39.2\% | 28.2\% | 28.9\%\\
Risk-free interest rate at grant date | 1.9\% | 2.9\% | 4.5\%\\
\textbf{Question:} what was the percent of the change of the expected volatility from 2008 to 2009?  
}

\gptbox{blue}{
\textbf{Gold Program:} subtract(39.2, 28.2), divide(\#0, 28.2) \\ 
\textbf{Gold Answer:} 0.39007 
}

\gptbox{violet}{
\textbf{ZS-FinDSL Reasoning Extraction Prompt Response:}\\
The percent of the change of the expected volatility from 2008 to 2009 can be calculated by subtracting the 2008 expected volatility from the 2009 expected volatility and then dividing the result by the 2008 expected volatility. \\ 
Step 1: Subtract the 2008 expected volatility from the 2009 expected volatility
2009 expected volatility - 2008 expected volatility = 39.2\% - 28.2\% = 11\%\\
Step 2: Divide the result by the 2008 expected volatility
11\% / 28.2\% = 0.39 or 39\%\\
Therefore, the percent of the change of the expected volatility from 2008 to 2009 is 39\%.\\
\textbf{ZS-FinDSL Program Extraction Prompt Response:}\\
\{"Program": \{"\#0":\{operation:"subtract", arg1:"39.2", arg2:"28.2"\},\\ 
\phantom{\{"Program": } "\#1":\{operation:"divide", arg1:"\#0", arg2:"28.2"\},\\ 
\phantom{\{} "Answer": "39\%"\}\\
\textbf{ZS-FinDSL Program:} subtract(39.2, 28.2), divide(\#0, 28.2)\\
\textbf{ZS-FinDSL Executed Answer:} 0.39007 
}
}; 
\end{tikzpicture}
\end{center}
\caption{An example where ZS-FinDSL extracted correct reasoning and successfully generated the DSL program. \label{fig:fin_dsl_2}}
\end{figure*}

\begin{figure*}[htb]
\begin{center}
\begin{tikzpicture}[node distance=12mm]
	\node (n1) [example_prompt] {
\qaid{HOLX/2009/page\_151.pdf-1}

\qabox{gray}{
\textbf{Passage:} table of contents hologic , inc . notes to consolidated financial statements ( continued ) ( in thousands , except per share data ) a summary of the company 2019s restricted stock units activity during $\ldots$ \\
Non-vested Shares | Number of Shares | Weighted-Average Grant-Date Fair Value\\
Non-vested at September 27, 2008 | 1,461 | \$31.23\\
Non-vested at September 26, 2009 | 2,770 | \$21.96\\
\textbf{Question:} by what percentage did the average price of wti crude oil increase from 2011 to 2013?  
}

\gptbox{blue}{
\textbf{Gold Program:} multiply(2770, 21.96)\\ 
\textbf{Gold Answer:} 60829.2 
}

\gptbox{deepblue}{
\textbf{ZS-FinPYT Program:}\\
non\_vested\_shares = 2770\\
weighted\_average\_grant\_date\_fair\_value = 21.96\\
ans = non\_vested\_shares * weighted\_average\_grant\_date\_fair\_value\\
print(ans) \# prints 60,532.2\\
\textbf{ZS-FinPYT Executed Answer:} 60829.2
}
}; 
\end{tikzpicture}
\end{center}
\caption{An example of successful Python program generation by ZS-FinPYT. \label{fig:fin_pyt_1}}
\end{figure*}

\begin{figure*}[htb]
\begin{center}
\begin{tikzpicture}[node distance=12mm]
	\node (n1) [example_prompt] {
\qaid{GS/2015/page\_188.pdf-4}

\qabox{gray}{
\textbf{Passage:} the goldman sachs group , inc . and subsidiaries notes to consolidated financial statements commercial lending . the $\ldots$ \\
of these amounts , \$ 2.86 billion and \$ 2.87 billion as of december 2015 and december 2014 , respectively , $\ldots$ \\
in millions | As of December 2015\\
$\ldots$ \\
2021 - thereafter | 1,160\\
Total | \$2,575\\
\textbf{Question:} in billions , what was the total for 2015 and 2014 relating to commitments to invest in funds managed by the firm?  
}

\gptbox{blue}{
\textbf{Gold Program:} add(2.86, 2.87) \\ 
\textbf{Gold Answer:} 5.73 
}

\gptbox{deepblue}{
\textbf{ZS-FinPYT Program:}\\
total\_2015 = 2.86 \#in billions\\
total\_2014 = 2.87 \#in billions\\
ans = total\_2015 + total\_2014 \#in billions\\
\textbf{ZS-FinPYT Executed Answer:} 5.73 
}
}; 
\end{tikzpicture}
\end{center}
\caption{An example of successful Python program generation by ZS-FinPYT. \label{fig:fin_pyt_2}}
\end{figure*}

In most of the cases, the LLM answering prompt of ZS-STD (text-davinci-003) generates only a value or a sentence, see Figure \ref{fig:fin_std_1} and Figure \ref{fig:fin_std_2} for details. In some cases, the answer extraction step fails as shown in Figure \ref{fig:std-ans-ext}. 
\begin{figure*}[htb]
\begin{center}
\begin{tikzpicture}[node distance=12mm]
	\node (n1) [example_prompt] {
\qaid{MRO/2013/page\_39.pdf-3}

\qabox{gray}{
\textbf{Passage:} iitem 7 . management 2019s discussion and analysis of financial condition and results of operations each of our segments is organized and managed based upon both geographic location and the nature $\ldots$ \\
Benchmark | 2013 | 2012 | 2011\\
WTI crude oil(Dollars per bbl) | \$98.05 | \$94.15 | \$95.11\\
Brent (Europe) crude oil(Dollars per bbl) | \$108.64 | \$111.65 | \$111.26\\
Henry Hub natural gas(Dollars per mmbtu)| \$3.65 | \$2.79 | \$4.04\\
\textbf{Question:} by what percentage did the average price of wti crude oil increase from 2011 to 2013?  
}

\gptbox{blue}{
\textbf{Gold Program:} subtract(98.05, 95.11), divide(\#0, 95.11)\\ 
\textbf{Gold Answer:} 0.03091
}

\gptbox{darkgreen}{
\textbf{ZS-STD LLM Answering Prompt Response:} 3.9\% \\
\textbf{ZS-STD Extracted Answer:} 3.9
}
}; 
\end{tikzpicture}
\end{center}
\caption{An example where ZS-STD's LLM answering prompt generates one value response.\label{fig:fin_std_1}}
\end{figure*}

\begin{figure*}[htb]
\begin{center}
\begin{tikzpicture}[node distance=12mm]
	\node (n1) [example_prompt] {
\qaid{UNP/2007/page\_25.pdf-4}

\qabox{gray}{
\textbf{Passage:} 2022 fuel prices 2013 crude oil prices increased at a steady rate in 2007 $\ldots$ \\
Millions of Dollars | 2007 | 2006 | 2005\\
Dividends paid | (364) | (322) | (314)\\
Free cash flow | \$487 | \$516 | \$234\\
\textbf{Question:} what was change in millions of free cash flow from 2005 to 2006?  
}

\gptbox{blue}{
\textbf{Gold Program:} subtract(516, 234) \\ 
\textbf{Gold Answer:} 282.0 
}

\gptbox{darkgreen}{
\textbf{ZS-STD LLM Answering Prompt Response:} The change in millions of free cash flow from 2005 to 2006 was an increase of \$282 million. \\ 
\textbf{ZS-STD Extracted Answer:} 282.0 
}
}; 
\end{tikzpicture}
\end{center}
\caption{An example where ZS-STD's LLM answering prompt generates one sentence response. \label{fig:fin_std_2}}
\end{figure*}

\begin{figure*}[htb]
\begin{center}
\begin{tikzpicture}[node distance=12mm]
	\node (n1) [example_prompt] {
\qaid{HOLX/2009/page\_151.pdf-1}

\qabox{gray}{
\textbf{Passage:} table of contents hologic , inc . notes to consolidated financial statements ( continued ) ( in thousands , except per share data ) a summary of the company 2019s restricted stock units activity during $\ldots$ \\
Non-vested Shares | Number of Shares | Weighted-Average Grant-Date Fair Value\\
Non-vested at September 27, 2008 | 1,461 | \$31.23\\
Granted. | 1,669 | 14.46\\
Vested | (210) | 23.87\\
Forfeited | (150) | 23.44\\
Non-vested at September 26, 2009 | 2,770 | \$21.96\\
\textbf{Question:} by what percentage did the average price of wti crude oil increase from 2011 to 2013?  
}

\gptbox{blue}{
\textbf{Gold Program:} multiply(2770, 21.96) \\ 
\textbf{Gold Answer:} 60829.2 
}

\gptbox{darkgreen}{
\textbf{ZS-STD LLM Answering Prompt Response:} The total fair value of non-vested shares as of September 26, 2009 is \$59,812. \\
\textbf{ZS-STD Extracted Answer:} float  
}
}; 
\end{tikzpicture}
\end{center}
\caption{An example where answer extraction failed for ZS-STD. \label{fig:std-ans-ext}}
\end{figure*}

The LLM responses with ZS-CoT details out the reasoning behind the answers and shows significantly better performance than ZS-STD. However, arithmetic errors results into a substantial drop in performance for both ZS-STD prompt and ZS-CoT. Examples demonstrating arithmetic errors are shown in Figure \ref{fig:arithmetic-error_1} and Figure \ref{fig:arithmetic-error_2}.

\begin{figure*}[htb]
\begin{center}
\begin{tikzpicture}[node distance=12mm]
	\node (n1) [example_prompt] {
\qaid{FIS/2016/page\_45.pdf-3}

\qabox{gray}{
\textbf{Passage:} financial statements . as of december 31 , 2016 $\ldots$ \\
cash flows from operations cash flows from operations were \$ 1925 million , \$ 1131 million and \$ 1165 million in 2016 , 2015 and 2014 respectively. $\ldots$ \\
Type of Obligations | Total | Less than 1 Year | 1-3 Years | 3-5 Years | More than 5 Years\\
$\ldots$ \\
Total | \$14,429 | \$1,068 | \$2,712 | \$3,264 | \$7,385\\
\textbf{Question:} what was the percentage change in cash flows from operations from 2015 to 2016?  
}

\gptbox{blue}{
\textbf{Gold Program:} subtract(1925, 1131), divide(\#0, 1131) \\ 
\textbf{Gold Answer:} 0.70203 
}

\gptbox{darkgreen}{
\textbf{ZS-STD LLM Answering Prompt Response:} The cash flows from operations increased by 794 million from 2015 to 2016, which is a 69.7\% increase. \\
\textbf{ZS-STD Extracted Answer:} 69.7 
}

\gptbox{magenta}{
\textbf{ZS-CoT Reasoning Prompt Response:} Cash flows from operations in 2015 = \$1131 million, Cash flows from operations in 2016 = \$1925 million. Therefore, the percentage change in cash flows from operations from 2015 to 2016 is (1925-1131)/1131 = 70.1\%. \\
\textbf{ZS-CoT Extracted Answer:} 70.1 
}
}; 
\end{tikzpicture}
\end{center}
\caption{An example of arithmetic errors made by ZS-STD and ZS-CoT prompts. \label{fig:arithmetic-error_1}}
\end{figure*}

\begin{figure*}[htb]
\begin{center}
\begin{tikzpicture}[node distance=12mm]
	\node (n1) [example_prompt] {
\qaid{RSG/2018/page\_94.pdf-1}

\qabox{gray}{
\textbf{Passage:} republic services , inc . notes to consolidated financial statements 2014 ( continued ) high quality financial  $\ldots$ \\
-  | 2018 | 2017 | 2016\\
Balance at beginning of year | \$38.9 | \$44.0 | \$46.7\\
Additions charged to expense | 34.8 | 30.6 | 20.4\\
Accounts written-off | ( 39.4) | ( 35.7) | ( 23.1)\\
Balance at end of year | \$34.3 | \$38.9 | \$44.0\\
\textbf{Question:} as of december 31 , 2018 what was the percentage decline in the allowance for doubtful accounts?  
}

\gptbox{blue}{
\textbf{Gold Program:} subtract(34.3, 38.9), divide(\#0, 38.9) \\ 
\textbf{Gold Answer:} -0.11825 
}

\gptbox{darkgreen}{
\textbf{ZS-STD LLM Answering Prompt Response:} The percentage decline in the allowance for doubtful accounts as of December 31, 2018 was 11.3\% (34.3 - 38.9 / 38.9). \\
\textbf{ZS-STD Extracted Answer:} 11.3 
}

\gptbox{magenta}{
\textbf{ZS-CoT Reasoning Prompt Response:} The balance at the beginning of 2018 was \$38.9 million. The balance at the end of 2018 was \$34.3 million. Therefore, the percentage decline in the allowance for doubtful accounts is 12.2\% (\$38.9 - \$34.3 / \$38.9).\\
\textbf{ZS-CoT Extracted Answer:} 12.2 
}
}; 
\end{tikzpicture}
\end{center}
\caption{An example of arithmetic errors made by ZS-STD and ZS-CoT prompts.\label{fig:arithmetic-error_2}}
\end{figure*}

The ZS-FinDSL performance is lower than ZS-FinPYT because the program extraction step fails for some cases where the reasoning step is correct as shown in Figure \ref{fig:program-error1} and Figure \ref{fig:program-error2}.
\begin{figure*}
\begin{center}
\begin{tikzpicture}[node distance=12mm]
	\node (n1) [example_prompt] {
\qaid{UNP/2007/page\_25.pdf-4}

\qabox{gray}{
\textbf{Passage:} 2022 fuel prices 2013 crude oil prices increased at a steady rate in 2007 $\ldots$ \\
Millions of Dollars | 2007 | 2006 | 2005\\
Dividends paid | (364) | (322) | (314)\\
Free cash flow | \$487 | \$516 | \$234\\
\textbf{Question:} what was change in millions of free cash flow from 2005 to 2006?  
}

\gptbox{blue}{
\textbf{Gold Program:} subtract(516, 234) \\ 
\textbf{Gold Answer:} 282.0 
}

\gptbox{violet}{
\textbf{ZS-FinDSL Reasoning:} The change in millions of free cash flow from 2005 to 2006 was an increase of \$282 million (\$516 million - \$234 million = \$282 million).\\
\textbf{ZS-FinDSL Program:}\\
\{"Program": \{"\#0":\{operation:"subtract", arg1:"516", arg2:"234"\},\\
\phantom{\{"Program": } "\#1":\{operation:"add", arg1:"\#0", arg2:"282"\},\\
\phantom{\{} "Answer": "282"\}\\
\textbf{ZS-FinDSL Executed Answer:} 564 
}
}; 
\end{tikzpicture}
\end{center}
\caption{An example where program extraction failed for ZS-FinDSL prompt. \label{fig:program-error1}}
\end{figure*}

\begin{figure*}[htb]
\begin{center}
\begin{tikzpicture}[node distance=12mm]
	\node (n1) [example_prompt] {
\qaid{ADBE/2008/page\_89.pdf-4}

\qabox{gray}{
\textbf{Passage:} summary fin 48 changes during fiscal 2008 , our aggregate changes in our total gross amount of unrecognized tax $\ldots$ \\
future tax returns , would result : \$ 57.7 million of unrecognized tax benefits would decrease the effective tax rate and \$ 68.9 million would decrease goodwill . $\ldots$\\
Beginning balance as of December 1, 2007 | \$201,808\\
Settlements with taxing authorities | (81,213)\\
Lapse of statute of limitations | (3,512) \\
\textbf{Question:} if the total fin 48 gross liability for unrecognized tax benefits at november 28, 2008 were recognized in the future, is the amount of unrecognized tax benefits that would decrease the effective tax rate greater than the amount that would decrease goodwill?  
}

\gptbox{blue}{
\textbf{Gold Program:} greater(57.7, 68.9) \\ 
\textbf{Gold Answer:} no 
}

\gptbox{violet}{
\textbf{ZS-FinDSL Reasoning:} The amount of unrecognized tax benefits that would decrease the effective tax rate is \$57.7 million. The amount that would decrease goodwill is \$68.9 million. Therefore, the amount of unrecognized tax benefits that would decrease the effective tax rate is less than the amount that would decrease goodwill.",
\\
\textbf{ZS-FinDSL Program:}\\
\{"Program": \{"\#0":\{operation:"subtract", arg1:"68.9", arg2:"57.7"\}, \\
\phantom{\{} "Answer": "False"\}\\
\textbf{ZS-FinDSL Executed Answer:} 11.2 
}
}; 
\end{tikzpicture}
\end{center}
\caption{An example where program extraction failed for ZS-FinDSL prompt. \label{fig:program-error2}}
\end{figure*}

\end{document}